\documentclass[letterpaper]{article} 
\usepackage{aaai25}  
\usepackage{times}  
\usepackage{helvet}  
\usepackage{courier}  
\usepackage[hyphens]{url}  
\usepackage{graphicx} 
\usepackage{natbib}  
\usepackage{caption} 
\usepackage{algorithm}
\usepackage{algorithmic}
\usepackage{booktabs}
\usepackage{newfloat}
\usepackage{listings}
\usepackage{amssymb}
\usepackage{comment}
\usepackage{makecell}
\usepackage{xcolor}
\usepackage{multicol}
\usepackage{multirow}
\usepackage{cleveref}
\DeclareCaptionStyle{ruled}{labelfont=normalfont,labelsep=colon,strut=off} 
\floatstyle{ruled}
\newfloat{listing}{tb}{lst}{}
\floatname{listing}{Listing}
\title{SGC-VQGAN: Towards Complex Scene
Representation via Semantic Guided Clustering
Codebook}
\author {
    Chenjing Ding \textsuperscript{\rm 1}, Chiyu Wang\textsuperscript{\rm 1}, Boshi Liu\textsuperscript{\rm 1}, Xi Guo\textsuperscript{\rm 1}, Weixuan Tang\textsuperscript{\rm 1}, Wei Wu\textsuperscript{\rm 1,2}\thanks{Corresponding author.}
}

\affiliations{
    \textsuperscript{\rm 1} Sensetime Research  \textsuperscript{\rm 2} Tsinghua University\\
    \{dingchenjing, liuboshi, guoxi, wuwei\}@sensetime.com, \{wangchiyu, tangweixuan\}@senseauto.com
}
\usepackage{bibentry}

\begin{document}

\maketitle

\begin{abstract}
Vector quantization (\textbf{VQ}) is a method for deterministically learning features through discrete codebook representations. Recent works have utilized visual tokenizers to discretize visual regions for self-supervised representation learning. However, a notable limitation of these tokenizers is lack of semantics, as they are derived solely from the pretext task of reconstructing raw image pixels in an auto-encoder paradigm. Additionally, issues like imbalanced codebook distribution and codebook collapse can adversely impact performance due to inefficient codebook utilization.
To address these challenges, We introduce SGC-VQGAN through Semantic Online Clustering method to enhance token semantics through Consistent Semantic Learning. Utilizing inference results from segmentation model , our approach constructs a temporospatially consistent semantic codebook, addressing issues of codebook collapse and imbalanced token semantics. Our proposed Pyramid Feature Learning pipeline integrates multi-level features to capture both image details and semantics simultaneously. As a result, SGC-VQGAN achieves SOTA performance in both reconstruction quality and various downstream tasks. Its simplicity, requiring no additional parameter learning, enables its direct application in downstream tasks, presenting significant potential.
\end{abstract}
\section{Introduction}
Vector quantization (VQ) is a technique used to learn features with discrete codebook representations, often implemented with a variational autoencoding model called VQ-VAE\cite{VQ-VAE}. It is commonly employed alongside deep generative modeling for tasks such as denoising diffusion probabilistic models and autoregressive models\cite{vqgan,gaia1, worlddreamer, VAM}. These methods involve a two-step process: first, learning a compressed discrete representation for each image, and then learning the prior distribution in the discrete latent space. This compression of redundant information greatly aids in training deep generative models on large-scale data. It's important to note that regardless of the performance of the prior model, the compression performance of VQ ultimately limits the overall generation performance.

\begin{figure}[t]
  \centering
  \includegraphics[width=1.0\linewidth]{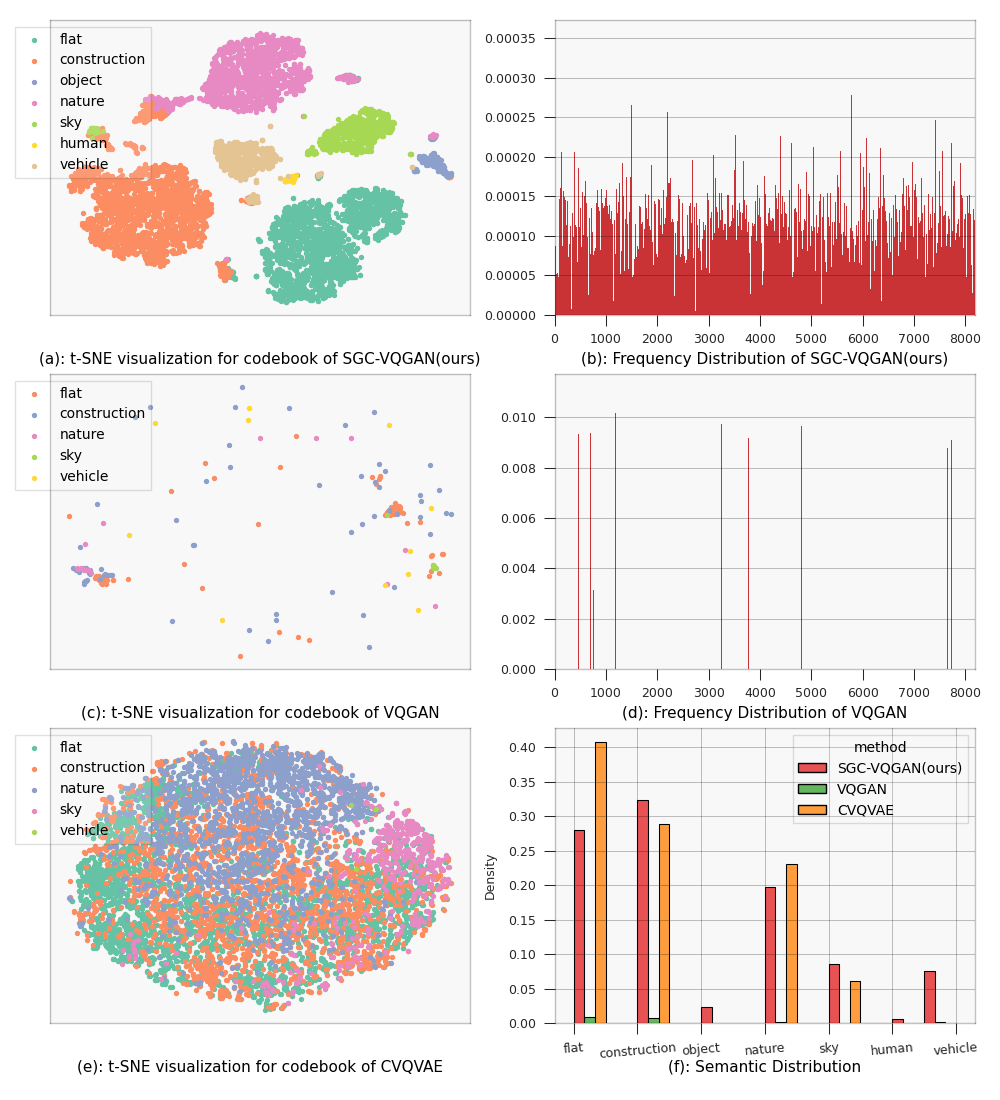}
  \caption{Semantic Performance Comparison; (a) tsne visualize for SGC-VQGAN(ours); Our method successfully enhanced codebook semantic; (b) codebook usage of SGC-VQGAN(ours); Our method achieve balanced and efficient utilization of codebook with $100\%$ active tokens; (c) tsne visualize for VQGAN\cite{vqgan}; Most tokens are not used and the other tokens are with poor clustering; (d) codebook usage of VQGAN; few tokens are active ($<10\%$); (e) tsne visualize for CVQ-VAE\cite{zheng2023online}; Most tokens are used, however these tokens are lack semantic significance; (f) semantic performance; Other methods lack specific semantic tokens, such as those representing human and vehicles, which are crucial for generating real-world scenarios. Our method successfully cover these classes. }
  \label{fig:vis_dist}
\end{figure}

Despite the achievements of VQ-VAE \cite{VQ-VAE} and VQGAN \cite{vqgan}, they face limitations when employed for complex scene generation tasks that demand high resolution, emphasize real-world dynamic scenarios, incorporate abundant contextual semantics, and take into account temporal aspects. \textbf{Firstly}, one prominent drawback is their lack of semantic understanding, stemming from their reliance solely on the auto-encoder approach for reconstructing raw image pixels. GAIA-1\cite{gaia1} suggests that enhancing the world model's performance hinges on directing the compression process towards meaningful representations, such as semantics, rather than solely prioritizing high-frequency signals. To address this, they employ the DINOv2\cite{oquab2023dinov2} model to guide compression. However, the semantic features in DINOv2 lack consistency across both time and space, as depicted in pipeline. Consequently, this method fails to consistently improve the semantic aspect of the codebook. \textbf{Secondly}, imbalanced codebook distribution and codebook collaspe can adversely impact performance due to inefficient codebook utilization;  Codebook collapse refers to the situation where most tokens become inactive and are not used during the generation stage. \cite{williams2020hierarchical} aims to reset infrequently used or unused entries by leveraging frequently accessed ones, while CVQ-VAE \cite{zheng2023online} proposes an online clustering algorithm that updates codebook entries using associated encoded features during the training stage. However, none of these approaches adequately address the imbalance in codebook utilization concerning semantics, so that the codebook are not fully utilized. As illustrated in Figure~\ref{fig:vis_dist}, we present a visualization of the semantic distribution of the codebook in VQGAN\cite{vqgan} and its improved versions\cite{zheng2023online} on the Nuscenes dataset. VQGAN\cite{vqgan} and CVQ-VAE\cite{zheng2023online} exhibits an alarming $0.7\%$ and $23.5\%$ in terms of semantic uniqueness respectively, indicating a lack of specific semantics for the majority of these tokens. For more details about semantic performance, please refer to Appendix and Table~\ref{tab:results}.

\begin{figure*}
  \centering
  \includegraphics[width=0.8\linewidth]{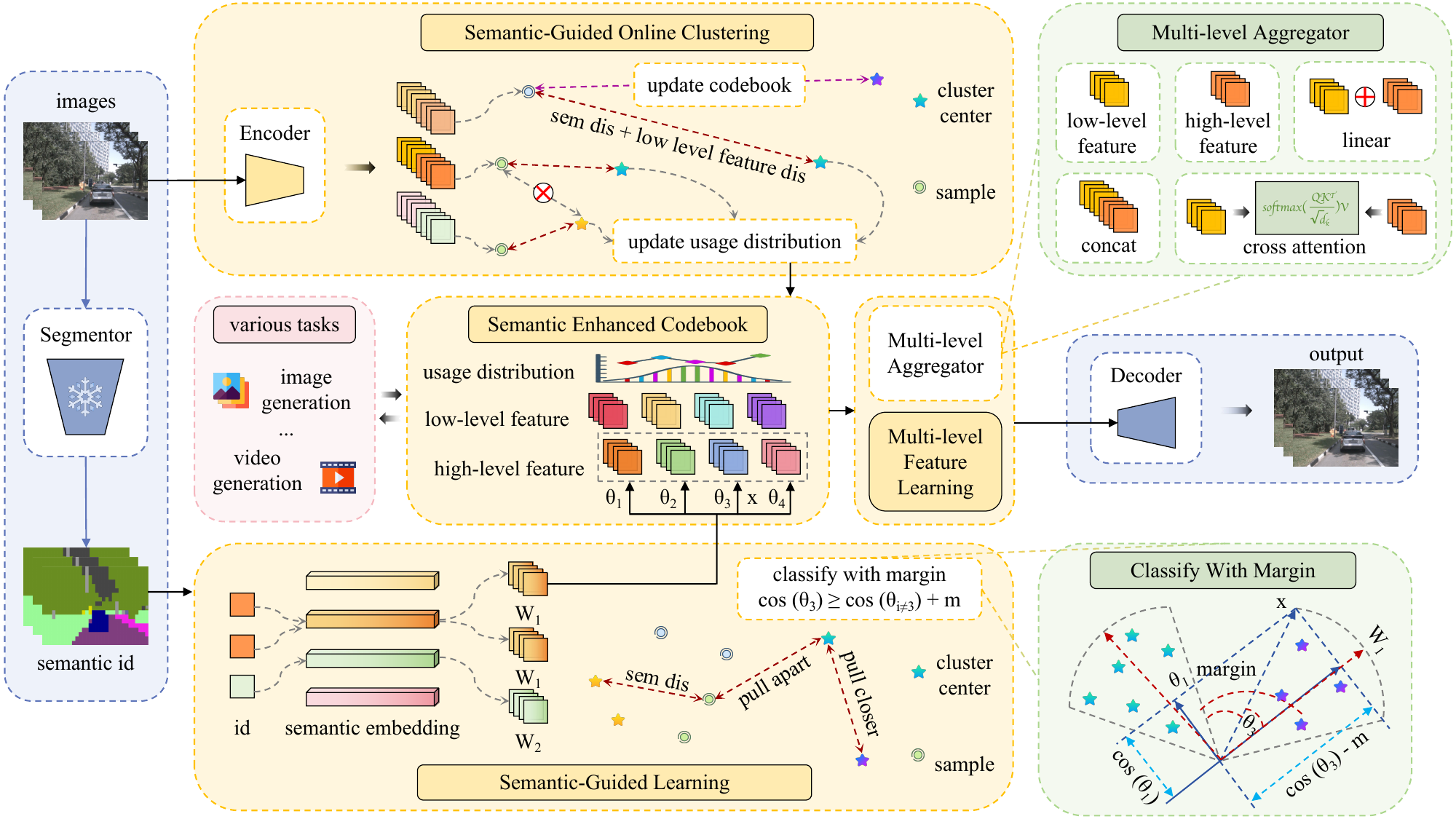}
  \caption{We introduce a Semantic Online Clustering method to enhance token semantics through Consistent Semantic Learning; Utilizing segmentation model inference results, our approach constructs a temporospatially consistent semantic codebook, addressing issues of codebook collapse and imbalanced token semantics;We use Pyramid Feature Learning pipeline integrates multi-level features to capture both image details and semantics simultaneously. Its simplicity allows for direct application in downstream tasks, offering significant potential.}
  \label{fig:pipeline}
\end{figure*}

To address these challenges, we propose the \textbf{Consistent Semantic Learning} approach, which leverages contrastive learning to guide compression while incorporating semantics. By leveraging the inference results from the segmentation model\cite{xie2021segformer}, our approach achieves temporospatially consistent semantic learning, as the semantic IDs remain consistent across both time and space. We formulate the compression task as a \textbf{Semantic Online Clustering} problem. This involves integrating high-level features into the codebook and computing multilevel distances between tokens and encoded features. These distances are then aggregated through a weighted sum to determine the clustering criteria for the codebook and encoded features. As the encoded features within the deep network undergo updates during training rather than remaining fixed, we apply a dynamic updating strategy to update multi-level features in the codebook. Additionally, our \textbf{Multi-level Feature Learning} pipeline combines low-level and high-level features to enhance both reconstruction performance and semantic understanding. Given that the decoder of VQGAN\cite{vqgan} is often not utilized in the second stage in complex scene generation applications\cite{gaia1, worlddreamer}, our pyramid feature learning approach focuses on simplicity without requiring additional parameter learning. This enables direct utilization of our method in downstream tasks, thereby maximizing its potential. The detailed process is demonstrated in Figure~\ref{fig:pipeline}.

In summary, our main contributions are as follows:
\begin{enumerate}
    \item We introduce a \textbf{Semantic Online Clustering} method to enhance token semantics through \textbf{Consistent Semantic Learning} without requiring highly accurate annotations. By utilizing segmentation model inference results, our approach creates a temporospatially consistent semantic codebook, addressing codebook collapse and imbalanced token semantics which improves performance in various tasks leveraging VQ as a prior model.
    \item The proposed \textbf{Pyramid Feature Learning} pipeline integrates multi-level features to simultaneously capture image details and semantics. Its simplicity, without requiring additional parameter learning, allows direct application in downstream tasks, offering significant potential.
    \item Our method undergoes extensive validation across diverse datasets and tasks to ascertain its performance.
\end{enumerate}

\section{Method}
\label{headings}

\subsection{Preliminary}

\paragraph{VQ-VAE}

VQ-VAE embeds a high-dimensional image ($x \in \mathbb{R}^{H \times W \times 3}$) with low-dimensional code vectors ($c_q \in \mathbb{R}^{h \times w \times n_q}$), where $n_q$ is the dimensionality of the codebook vectors. The feature tensor is compactly represented as $h \times w$ indices. This compact representation is achieved via training a reconstruction task $\hat{x} = G_\theta(c_q) = G_\theta(q(\hat{z})) = G_\theta(q(E_\phi(x)))$, where $E_\phi$ and $G_\theta$ are the encoder and decoder, $q$ and $\hat{z}$ denotes quantization and encoded features. During training, the encoder $E_\phi$, decoder $G_\theta$, and codebook are jointly optimized by minimizing the following objective:
\begin{equation} 
\label{eq:N_k}
    loss = \Vert x - \hat{x} \Vert_2  +  \Vert sg[E_\phi(x)] - c_q \Vert_2 + \gamma * \Vert E_\phi(x) - sg[c_q] \Vert_2
\end{equation}
where $sg$ denotes a stop-gradient operator, and $\gamma$ is the weight of the last term commitment loss.
 
\paragraph{Online Clustering}
VQ-VAE \cite{VQ-VAE} suffers from codebook collapse, where only a few codevectors receive useful gradients, leaving most unused. CVQGAN \cite{zheng2023online} addresses this by using an online clustering approach to update unused codevectors. It calculates the average usage of codevectors in each training mini-batch and updates their counts accordingly as Eq~\ref{eq:N_k}. Anchors are selected from encoded features to modify less-used or unused codevectors more significantly. Decay values are computed based on average usage as Eq~\ref{eq:a_k}, and codebook features are updated by combining previous values with sampled anchors, weighted by the decay values as Eq~\ref{eq:c_k}.
\begin{equation} 
\label{eq:N_k}
    N^{t}_{k} =   N^{t - 1}_{k} * \gamma + \frac{n^t_k}{Bhw}*(1 - \gamma),
\end{equation}
\begin{equation} 
\label{eq:a_k}
    a^t_k = exp^{-N^t_k * K * \frac{10}{1 - \gamma} - \epsilon},
\end{equation}
\begin{equation} 
\label{eq:z_k}
    \hat{e}_{k}^{t} = min_i(\Vert{e_{i}^t - c_k^t} \Vert_2),
\end{equation}
\begin{equation}
\label{eq:c_k}
    c_k^t = c_k^{t - 1} * ( 1 - a^t_k) + \hat{e}_{k}^{t} * a^t_k,
\end{equation}
where $N^{t}_{k}$ is the number of encoded features in a mini-batch, $Bhw$ denotes the number of features on Batch, height and width. $\gamma$ is a decay hyperparameter, $\epsilon$ is a small constant.
\subsection{Semantic Online Clustering}
\label{sec:Semantic Online Clustering}
CVQ-VAE\cite{zheng2023online} proposes an online clustering algorithm that updates codebook entries using associated encoded features during training. However, it relies only on pixel-level elements and lacks high-level semantic features. To address this, we incorporate semantic embedding and semantic distance into the online clustering process.

\textbf{Firstly}, we use a segmentation model \cite{gao2022large, xie2021segformer} to get semantic ids of each image. Suppose there are $N$ classes in total, we will learn $N$ semantic embedding $ W_i$, where $i\in[0, N-1]$; \textbf{Then}, we incorporate high-level features to codebook and compute multilevel distances of encoded feature separately, using these distances to cluster the codebook and encoded features as Eq~\ref{eq:e_k_multi}; \textbf{Finally}, since the data points, the encoded features in the deep network, are also updated during training instead of being fixed, a dynamic updating strategy must account for these changing feature representations. Therefore, we use exponential moving average(EMA) approach with  $ W_{i} $ and encoded feature to update the codebook together as Eq~\ref{eq:c_k_h} and Eq~\ref{eq:c_k_l}. As a result, along with the training stage moves on and the hit frequency of codebook entries gets larger, $ W_{i} $ will dominate the updating of high-level features of the codebook than encoded features resulting in more coherent and unified high-level features. Leveraging semantic embedding, the codebook entries become highly discriminative, each aligning with a specific semantic class. This approach minimizes semantic ambiguity, prevents the reuse of identical codebook entries across different classes, and ensures efficient utilization across all classes.
\begin{equation}
\label{eq:e_k_multi}
    \hat{e}_{k}^{t} = min_i(\Vert{\tilde{e}_{i}^t - \tilde{c}_{k}^t} \Vert_2 + \beta * \Vert{\widehat{e}_{i}^t - \widehat{c}_{k}^t} \Vert_2),
\end{equation}
\begin{equation}
\label{eq:c_k_h}
    \widehat{c}_k^t = \widehat{c}_k^{t - 1} * ( 1 - a_k^t) + (\hat{e}_{k}^{t}{}_{high} * a_k^t + (W_i) * (1 - a_k^t)) * a_k^t,
\end{equation}
\begin{equation}
\label{eq:c_k_l}
    \tilde{c}_k^t = \tilde{c}_k^{t - 1} * ( 1 - a_k^t) + \hat{e}_{k}^{t}{}_{low} * a_k^t,
\end{equation}

$\tilde{e}$ and $\tilde{c}$ denote low level features of encoded features and codebook, while $\widehat{e}$ and $\widehat{c}$ denote high level features.

\subsection{Consistent Semantic Learning}
\label{sec:Consistent Semantic Learning}
GAIA-1\cite{gaia1} utilizes DINOv2\cite{oquab2023dinov2} to integrate high-level semantic features. However, the semantic features in DINOv2\cite{oquab2023dinov2} lack consistency across time and space. In contrast, we employ the semantic class that maintains temporal and spatial consistency to guide the learning process. Our method does not require extremely accurate semantic annotations. By leveraging the inference results of the segmentation model\cite{gao2022large, xie2021segformer}, our approach achieves temporospatially consistent semantic learning.

\textbf{Initially}, we get each token entry $c_q$ with its class according to its index from the inference results of the segmentation model; \textbf{Next}, we classify tokens with the same class as positive samples and with different classes as negative samples. \textbf{Subsequently}, similar to \cite{wang2018cosface}, we minimize the objective as Eq~\ref{eq:c_k_l} by fixing the magnitude of x to eliminate radial variations, the model learns separable features in the angular space. 
\begin{equation}
\label{eq:c_k_l}
    loss = -\frac{1}{N} \sum_i log\frac{e^{s*(cos(\theta_{y_i,i}) - m) }}{ e^{s*(cos(\theta_{y_i,i}) - m )}+\sum_{j\neq y_i} e^{s*cos(\theta_{j,i})} },
\end{equation}
s is the magnitude of the codebook sample with i-th codebook entry; m is the margin; $\theta_{y_i,i}$ is the angle between semantic embedding $W_{y_i}$ and i-th codebook entry with ${y_i}$ class label; Besides the above loss, we also implement the other popular contrastive learning loss\cite{qian2019softtriple, liu2017sphereface, liu2016large} for ablation study detailed in Seq.Ablation Experiments.

\begin{figure*}[t]
  \centering
  \includegraphics[width=1.0\linewidth]{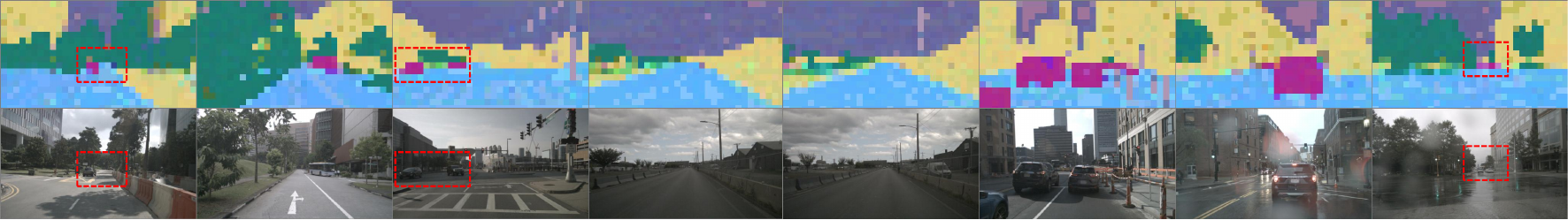}
  \caption{PCA visulaize for codebook; Up: we use pca to visualize our codebook; Bottom: the reference images.}
  \label{fig:pca}
\end{figure*}

\begin{figure}[t]
  \centering
  \includegraphics[width=1.0\linewidth]{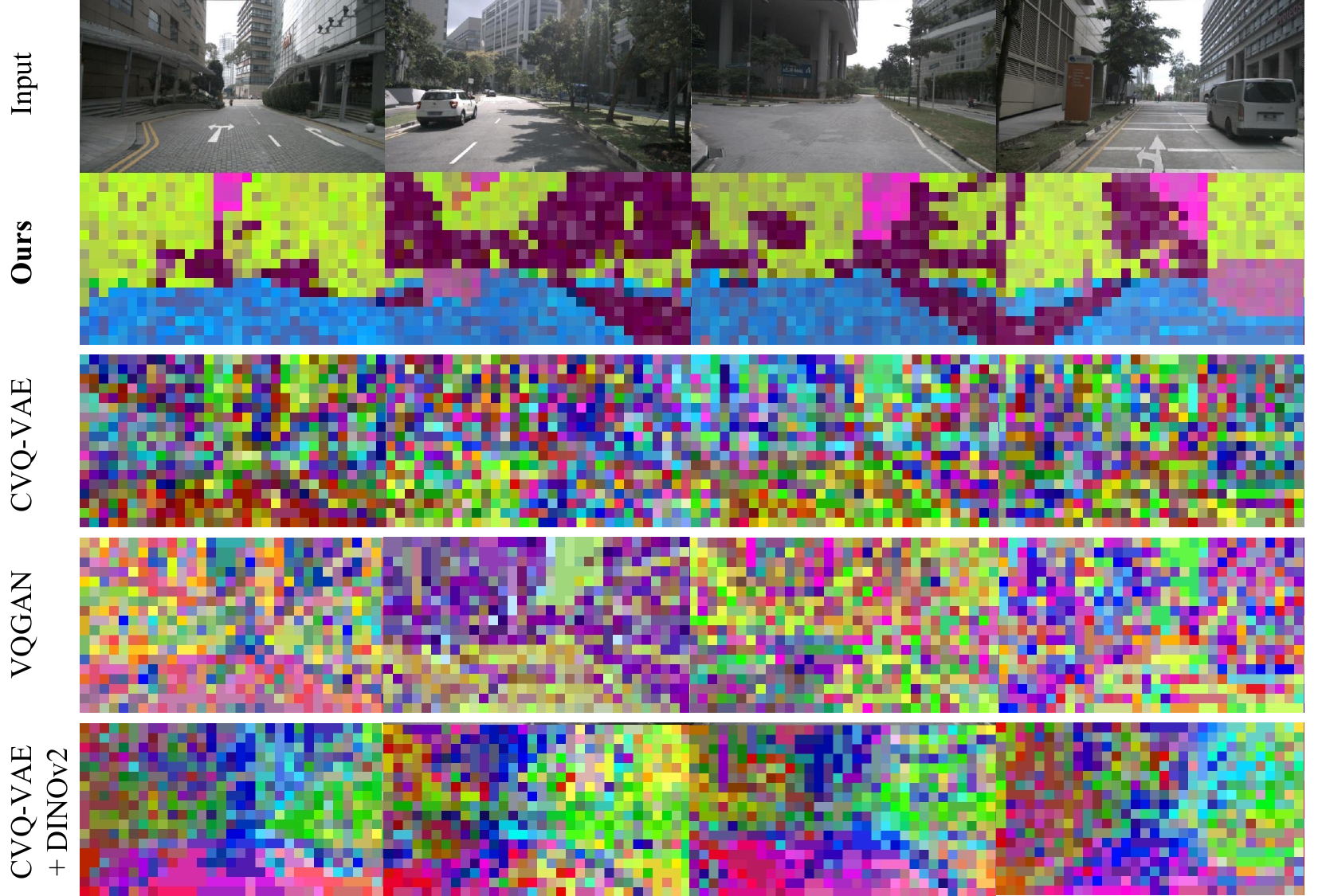}
  \caption{PCA visualization of the codebook for comparison across different methods.}
  \label{fig:mask_pca}
\end{figure}

\subsection{Multi-level Feature Learning}
\label{sec:Multi-level Feature Learning}
Semantic-guided features result in high-level, meaningful representations that suppress high-frequency signals. To capture both image details and semantics simultaneously, we introduce multi-level feature learning (\textbf{MFL}). 

\textbf{First}, we divide the codebook features into N different levels based on the intensity of semantic guidance, determined by the weight function detailed in Eq~\ref{eq:sigma} of the loss Eq~\ref{eq:c_k_l}. Specifically, Eq~\ref{eq:sigma} is monotonically decreasing function, the sum is 1 and $w(N)=0$; the lowest-level features are not guided by semantics at all.

\textbf{Second} Different level features are applied differently.

In \textbf{Sec.Semantic Online Clustering}, we use mixed features to cluster encoded features and the codebook. In \textbf{Sec.Consistent Semantic Learning}, high-level features are used to cluster the codebook itself, with contrastive learning guiding the compression process towards meaningful representations. Then, an aggregator merges the different level features as input to the decoder.

\textbf{Second}, features at different levels are utilized in distinct ways. In \textbf{Sec. Semantic Online Clustering}, a combination of features is employed to cluster the encoded features and the codebook, as the encoded features are updated online and guide the codebook preserving image's details. In \textbf{Sec. Consistent Semantic Learning}, only high-level features are used to directly cluster the codebookitself, minimizing the influence of RGB noise with contrastive learning guiding the compression process to produce meaningful representations. Then, an aggregator combines the features from various levels as input to the decoder.
\begin{equation}
\label{eq:sigma}
    w(l) = \frac{e^{-\alpha * l} - e^{-\alpha * N}}{\sum_{j = 1}^{N-1} (e^{-\alpha * j} - e^{-\alpha * N})},  1 \leq l \leq N;
\end{equation}
\textbf{Finally}, an aggregator merges the different level features as input to the decoder. There are five available modes: linear, concat, low-level-only, high-level-only, and cross-attention.
\textbf{Linear:} The different level features are combined through a weighted sum; 
\textbf{Concat:} The different level features are concatenated together; 
\textbf{Low-level-only:} Only the low-level features are sent to the decoder with the weight  $< \sigma$; 
\textbf{High-level-only:} Only the high-level features are sent to the decoder with the weight $> \sigma$; 
\textbf{Cross attention:} We use low-level features as query, and do cross attention with high-level features; 

In Sec. Experiment, we demonstrate that even without any additional parameters, our multi-level features perform the best in terms of image quality and semantic representations; In all, our multi-level feature can learn image details and semantics simultaneously, eliminating the need for additional parameter learning. This allows for direct integration into downstream tasks, thereby it greatly enhances the potential for significantly expanding the range of applications.

\subsection{Modeling Prior Distribution}
After embedding an image into a token sequence, we can estimate the underlying prior distribution over the discrete space, facilitating unconditional image, high-resolution image and video generation. As our primary objective is to enhance the codebook learning stage, we utilize the existing setup for the subsequent stage directly.
\label{sec:Modeling Prior Distribution}
\paragraph{Image Generation} 
After embedding an image ( I ) into a token sequence $s = \{s_1, · · ·, s_{h * w} \} $, the image generation process can be naturally formulated as an autoregressive next-symbol prediction problem. Specifically, we maximize the likelihood function.
\begin{equation}
\label{eq:p(s)}
    p(s) = \prod_i{p(s_i|s_{< i})},
\end{equation}
where $ p(s_i|s_{<i})$ is the probability of having the next symbol $s_i$ given all the previous symbols $s_{<i}$.

\paragraph{Video Prediction} 
Video prediction can be formulated using an autoregressive transformer network that models the sequential input. The training objective is to predict the next image token in the sequence, conditioned on all past tokens, using causal masking in the attention matrix of the transformer blocks. Then, we minimize the following objective:
\begin{equation}
\label{eq:p(s)}
    loss = - \sum_{t = 1} ^T  \sum_{i = 1}^n {log p(s_{t, i}|s_{< i}, z_{t, j < i})}.
\end{equation}

\section{Experiments}
\subsection{Experimental Details}
\textbf{Datasets.}
We evaluated the proposed method on multiple datasets, including ImageNet \cite{deng2009imagenet}, COCO \cite{lin2014microsoft}, and Nuscenes \cite{caesar2020nuscenes}. We excluded MNIST \cite{lecun1998gradient} and CIFAR10 \cite{krizhevsky2009learning} due to their limited diversity and simplistic content. We instantiated SGC-VQGAN for unconditional image generation tasks with $256 \times 256$ images from ImageNet and COCO, consistent with the original VQGAN\cite{vqgan}. Additionally, we applied SGC-VQGAN to both unconditional image and video generation tasks using the Nuscenes dataset, setting the image size to $288 \times 512$ to maintain the aspect ratio. We use Segformer \cite{xie2021segformer} and LUSS \cite{gao2022large} to infer segmentation results of Nuscenes and ImageNet\cite{deng2009imagenet} dataset;

\begin{table}[t]
\centering
\resizebox{1.0\linewidth}{!}{
\begin{tabular}{c|c|c|c|c|c}
\toprule
\textbf{Model} & 0.1 & 0.3 & 0.5 & 0.7 & 0.9 \\
\midrule
VQGAN & 0.0258 & 0.0219 & 0.0076 & 0.0060 & 0.0052 \\ 
CVQ-VAE & 1.0000 & 0.9542 & 0.2351 & 0.0093 & 0.0000 \\ 
\midrule
\textbf{Ours} (dim=8)& \textbf{1.0000} & \textbf{0.9999} & \textbf{0.9674} & \textbf{0.8978} & \textbf{0.7511} \\ 
\bottomrule
\end{tabular}
}
\caption{Comparison of semantic uniqueness; We compute the hit numbers of codebooks and classify these hits according to pixel semantics. The counts are then used as probabilities for these semantic classes of every entry in codebook. Finally, we provide the ratio of probabilities that exceed a given threshold. For more details, please refer to Appendix, Section Semantic Performance Analysis.}
\label{tab:results}
\end{table}

\textbf{Metrics.}
We evaluated the quality of the proposed methods on three tasks: image reconstruction , image generation , and video generation. \textbf{For image reconstruction task}, we used traditional patch-level evaluation metrics, including Peak Signal-to-Noise Ratio (PSNR), Structure Similarity Index (SSIM),  feature-level Learned Perceptual Image Patch Similarity (LPIPS)  \cite{lpips}.  \textbf{For image generation task}, we used one of the most common evaluation metric Fréchet Inception Distance (FID) \cite{heusel2017gans} to evaluate the quality and diversity of generated images. \textbf{For video generation task}, we used Fréchet Video Distance (FVD) \cite{unterthiner2019fvd}. In addition, we \textbf{evaluate codebook semantics} using the Silhouette Score (SS) \cite{rousseeuw1987silhouettes} and Davies-Bouldin Index (DBI) \cite{davies1979cluster}, where higher SS values and lower DBI values indicate better clustering quality which are widely used and effective metrics for evaluating clustering quality, characterized by them robustness and intuitiveness\cite{dbiapp3, ssapp1, dbiapp1, dbiapp2}. For more details, please refer to Appendix.

\definecolor{myLightBlue}{rgb}{0.3, 0.3, 1.0}
\definecolor{myLightRed}{rgb}{1.0, 0.3, 0.3}
\begin{table}[t]
\centering
\resizebox{1.0\linewidth}{!}{
\begin{tabular}{c|c|c|c|c|c|c|c}
\hline
\textbf{Model} & Latent Size & $\left| \mathcal{Z} \right|$  & PSNR$\uparrow$ & SSIM$\uparrow$ & LPIPS$\downarrow$  & SS$\uparrow$ & DBI$\downarrow$ \\ \hline
 
VQGAN\cite{vqgan}  & 18$\times$32 & 8192 & 26.24 & 0.7603 & 0.2836 & \textbf{\textcolor{myLightBlue}{-0.0081}} & \textbf{\textcolor{myLightBlue}{4.3211}} \\
CVQVAE\cite{zheng2023online} & 18$\times$32 & 8192 & \textbf{\textcolor{myLightRed}{27.37}} & \textbf{\textcolor{myLightRed}{0.8007}} & \textbf{\textcolor{myLightRed}{0.2248}} & -0.0297 & 4.7772 \\
\textbf{SGC-VQGAN} (Ours) & 18$\times$32 & 8192 & \textbf{\textcolor{myLightBlue}{26.78}} & \textbf{\textcolor{myLightBlue}{0.7946}} & \textbf{\textcolor{myLightBlue}{0.2395}} & \textbf{\textcolor{myLightRed}{0.7366}} & \textbf{\textcolor{myLightRed}{0.4455}} \\ \hline

\end{tabular}
}
\caption{Quantitative reconstruction results on the validation splits of Nuscenes.  $\left| \mathcal{Z} \right|$ denotes the size of codebook. (Red indicates the best, while Blue indicates the second-best.)}
\label{tab:quatitative recon results Nus}
\end{table}
\definecolor{myLightBlue}{rgb}{0.3, 0.3, 1.0}
\definecolor{myLightRed}{rgb}{1.0, 0.3, 0.3}
\begin{table}[t]
\centering
\resizebox{1.0\linewidth}{!}{
\begin{tabular}{c|c|c|c|c|c}
\hline
\textbf{Dataset}  & \textbf{Model} & Latent Size & $\left| \mathcal{Z} \right|$  & FID$\downarrow$ & FVD$\downarrow$  \\ \hline

\multirow{3}{*}{ImageNet} &  VQGAN          & 32$\times$32 & 16384 & 25.65 & - \\
                          & CVQVAE         & 32$\times$32 & 16384 & \textbf{\textcolor{myLightBlue}{24.87}} & - \\
                          & \textbf{SGC-VQGAN} (Ours) & 32$\times$32 & 16384 & \textbf{\textcolor{myLightRed}{19.92}} & - \\ \hline

\multirow{3}{*}{Nuscenes} &  VQGAN          & 18$\times$32 & 8192  & 24.01 & 1431.04 \\
                          & CVQVAE         & 18$\times$32 & 8192  & \textbf{\textcolor{myLightBlue}{20.72}} & \textbf{\textcolor{myLightBlue}{1372.72}} \\
                          & \textbf{SGC-VQGAN} (Ours) & 18$\times$32 & 8192  & \textbf{\textcolor{myLightRed}{19.40}} & \textbf{\textcolor{myLightRed}{825.95}} \\ \hline
\end{tabular}
}
\caption{Quantitative results for unconditional image generation on the validation splits of ImageNet\cite{deng2009imagenet} (50,000 images), COCO\cite{lin2014microsoft} (5,000 images) and Nuscenes\cite{caesar2020nuscenes}. Quantitative results for video pmyLightRediction on the validation splits of Nuscenes.  $\left| \mathcal{Z} \right|$ denotes the size of codebook. (Red indicates the best, while Blue indicates the second-best.)}
\label{tab:quatitative gen results}
\end{table}

\textbf{Network structures and implementation details.}
Our implementation is based on the officially released VQ-GAN architecture \cite{vqgan}. To ensure a fair comparison, we used consistent hyperparameters across all methods in our experiments. For the COCO and ImageNet datasets, images were downsampled by a factor of 8. For Nuscenes, images were resized from 288x512x3 to a grid of tokens with dimensions 18x32x32. We set the hyperparameters according to the baseline VQ-GAN work. In the first stage, we trained all models with a batch size of 10 across 8 Tesla A800 GPUs for 40 epochs focused on image reconstruction. In the second stage, for video generation and high-resolution image generation, we trained all models with a batch size of 2 across 8 Tesla A800 GPUs for 40 epochs.

\begin{figure*}[t]
  \centering
  \includegraphics[width=1.0\linewidth]{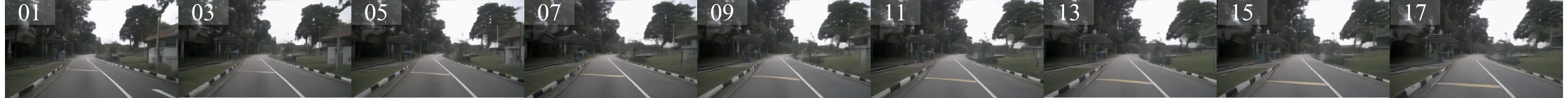}
  \caption{Video generation for complex scenarios. The numbers represent the video frame numbers.}
  \label{fig:video}
\end{figure*}

\begin{figure}[t]
  \centering
  \includegraphics[width=1.0\linewidth]{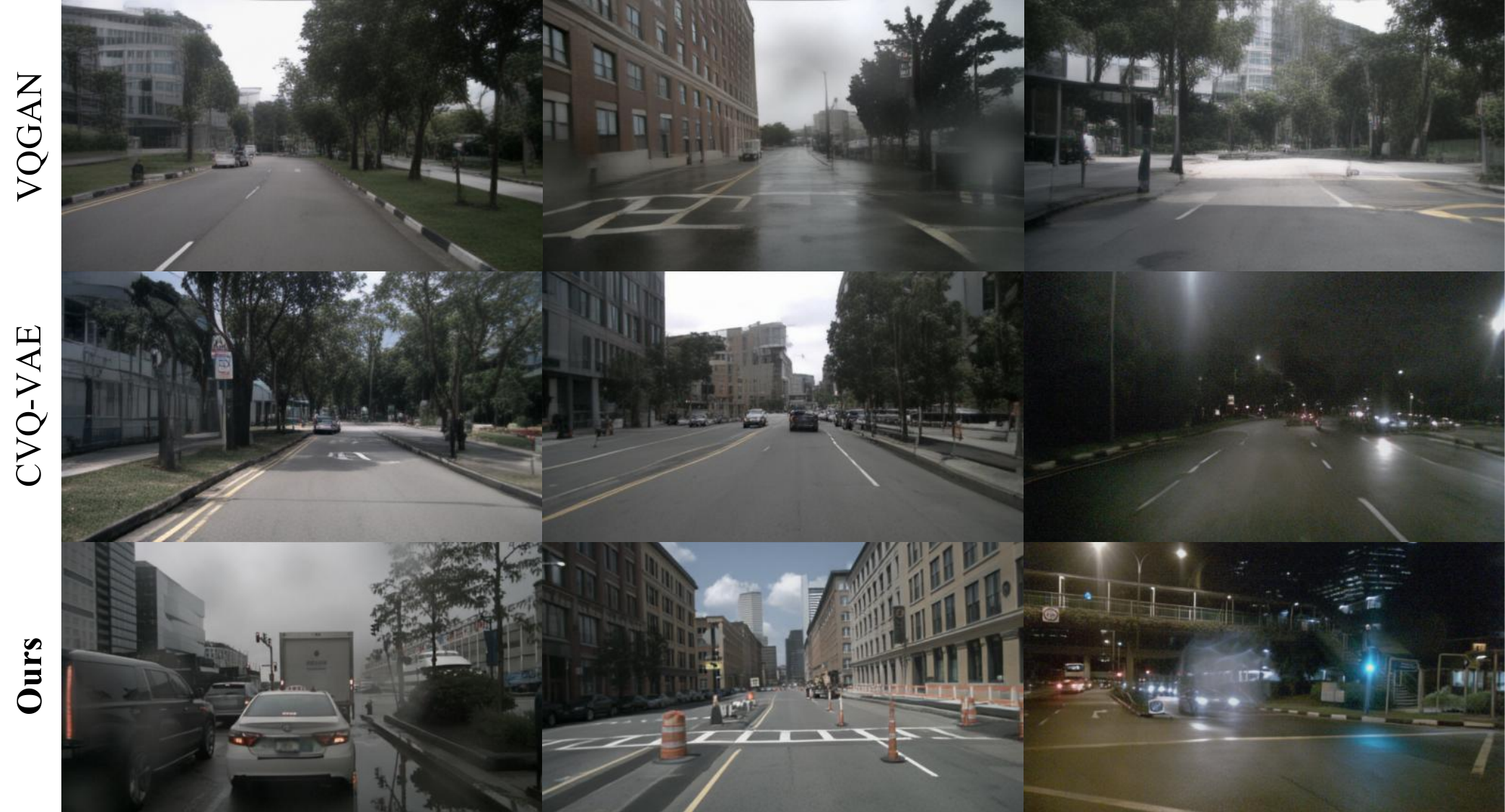}
  \caption{Image Generation in Nucenes dataset. Our model achieve highest image fidelity in terms of vehicles, zebra crossing and trees.}
  \label{fig:gen}
\end{figure}

\begin{figure}[htbp]
  \centering
  \includegraphics[width=1.0\linewidth]{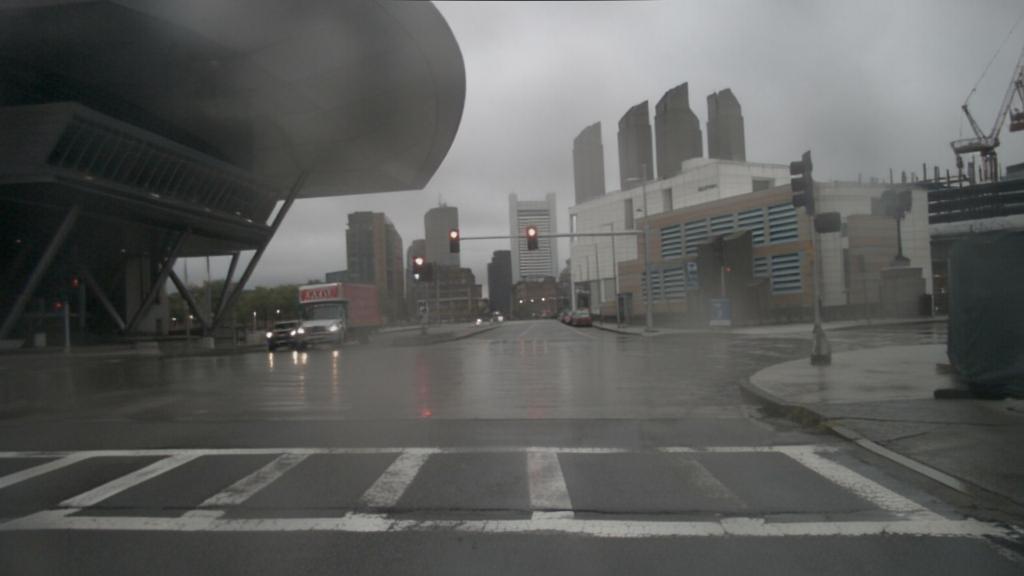}
  \caption{High resolution image generation for complex scenarios  with detailed and high-fidelity results(red lights and reflections).}
  \label{fig:high-re}
\end{figure}

\definecolor{myLightBlue}{rgb}{0.3, 0.3, 1.0}
\definecolor{myLightRed}{rgb}{1.0, 0.3, 0.3}
\begin{table}[t]
\centering
\resizebox{1.0\linewidth}{!}{
\begin{tabular}{c|c|c|c|c|c|c}
\hline                  
 \textbf{Dataset}         & \textbf{Model} &  Latent Size & $\left| \mathcal{Z} \right|$  & PSNR$\uparrow$ & SSIM$\uparrow$ & LPIPS$\downarrow$  \\ \hline
 \multirow{7}{*}{ImageNet} & VQGAN \cite{vqgan}  & 16$\times$16    & 1024    & 19.47 & 0.5214 & 0.1950  \\ 
                            & VQGAN \cite{vqgan} & 16$\times$16    & 16384    & 19.93 & 0.5424 & 0.1766  \\ 
                            &ViT-VQGAN \cite{yu2021vector}  & 32$\times$32    & 8192   & - &- & - \\ 
                             & RQ-VAE \cite{lee2022autoregressive} &  8$\times$8$\times$16  & 16384  & - & - & - \\  
                             &MoVQ \cite{zheng2022movq} & 16$\times$16$\times$4  & 1024  & 22.42 & 0.6731 &\textbf{\textcolor{myLightRed}{0.1132}} \\  
                             & CVQ-VAE \cite{zheng2023online} & 32$\times$32 & 16384  & \textbf{\textcolor{myLightBlue}{22.87}} & \textbf{\textcolor{myLightBlue}{0.6863}} & \textbf{\textcolor{myLightBlue}{0.1692}} \\
                             &\textbf{SGC-VQGAN} (Ours) & 32$\times$32 & 16384  & \textbf{\textcolor{myLightRed}{23.26}} & \textbf{\textcolor{myLightRed}{0.6903}} & 0.1745 \\ \hline
\multirow{3}{*}{COCO}       &  VQGAN \cite{vqgan} & 32$\times$32 & 16384  & 21.16 & 0.6188 & 0.2136 \\
                            & CVQGAN \cite{zheng2023online} & 32$\times$32 & 16384  & \textbf{\textcolor{myLightBlue}{22.48}} & \textbf{\textcolor{myLightRed}{0.6876}} & \textbf{\textcolor{myLightRed}{0.1660}} \\
                            & \textbf{SGC-VQGAN} (Ours) & 32$\times$32 & 16384  & \textbf{\textcolor{myLightRed}{22.65}} & \textbf{\textcolor{myLightBlue}{0.6818}} & \textbf{\textcolor{myLightBlue}{0.1749}} \\ \hline
\end{tabular}
}
\caption{Quantitative reconstruction results on the validation splits of ImageNet\cite{deng2009imagenet} (50,000 images), COCO\cite{lin2014microsoft} (5,000 images).  $\left| \mathcal{Z} \right|$ denotes the size of codebook. (Red indicates the best, while Blue indicates the second-best.)}
\label{tab:quatitative recon results}
\end{table}
\definecolor{myLightBlue}{rgb}{0.3, 0.3, 1.0}
\definecolor{myLightRed}{rgb}{1.0, 0.3, 0.3}
\begin{table}[t]
\centering
\resizebox{1.0\linewidth}{!}{
\begin{tabular}{c|c|c|c|c|c}
\hline                 
\textbf{Model} &  Latent Size & $\left| \mathcal{Z} \right|$  & PSNR$\uparrow$ & SSIM$\uparrow$ & LPIPS$\downarrow$  \\ \hline
CVQ-VAE \cite{zheng2023online} & 32$\times$32 & 16384 & 22.20 & 0.6760 & \textbf{\textcolor{myLightRed}{0.1730}} \\
\textbf{SGC-VQGAN} (Ours) & 32$\times$32 & 16384 & \textbf{\textcolor{myLightRed}{22.94}} & \textbf{\textcolor{myLightRed}{0.6803}} & 0.1738 \\ \hline
\end{tabular}
}
\caption{Quantitative reconstruction results on the validation splits of COCO\cite{lin2014microsoft} (5,000 images) from a model trained on ImageNet-S-919\cite{gao2022large}. $\left| \mathcal{Z} \right|$ denotes the size of codebook. ( Red indicates the best.)}
\label{tab:zeroshot}
\end{table}

\begin{figure*}[t]
  \centering
  \includegraphics[width=1.0\linewidth]{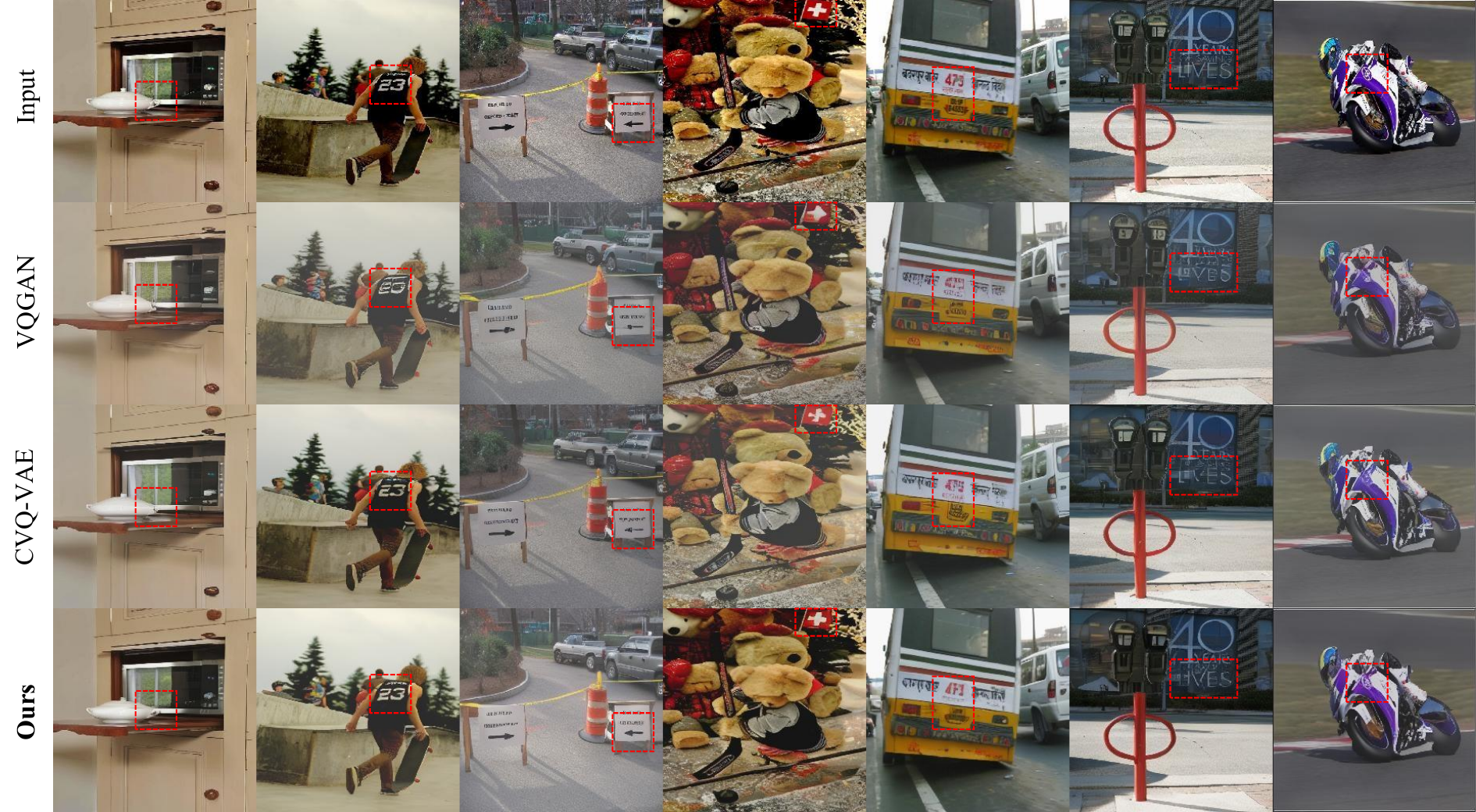}
  \caption{Reconstruction from different models on COCO. Compared to VQGAN and CVQ-VAE, our proposed model improves the reconstruction quality (highlighted in red boxes).}
  \label{fig:recons}
\end{figure*}

\begin{figure}[t]
  \centering
  \includegraphics[width=1.0\linewidth]{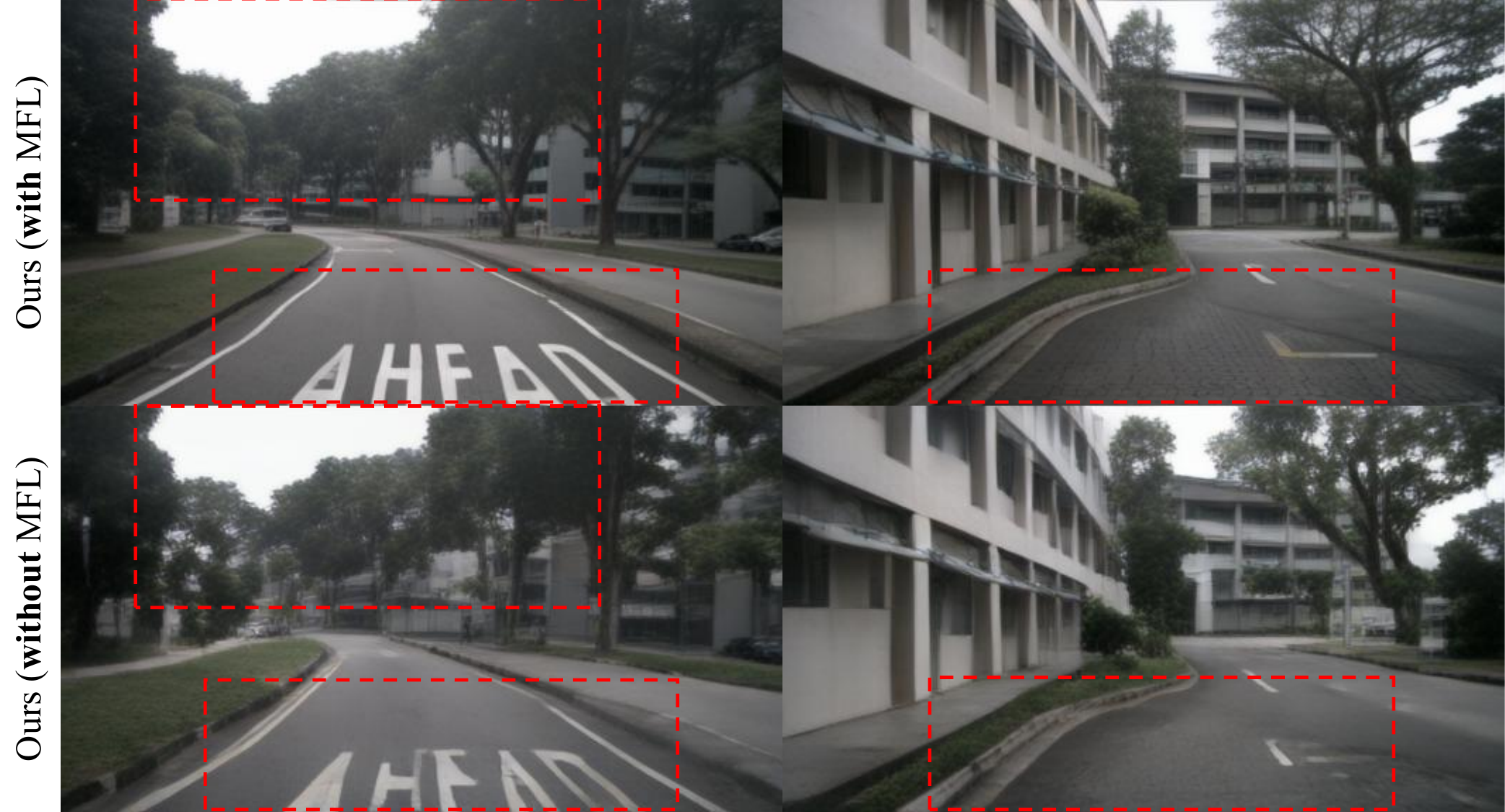}
  \caption{Comparison of reconstruction results between using multilevel features learning (MFL) method and not.}
  \label{fig:ablation_multilevel}
\end{figure}

\begin{table}[t]
\centering
\resizebox{1.0\linewidth}{!}{
\begin{tabular}{c|c|c|c|c|c|c}
\hline                  
 \textbf{Model} & Latent Size & $\left| \mathcal{Z} \right|$  & PSNR$\uparrow$ & SSIM$\uparrow$ & LPIPS$\downarrow$  & DBI$\downarrow$ \\ \hline
 
 SGC-VQGAN \textbf{wo MFL} & 18$\times$32 & 8192 & 23.1339 & 0.7207 & 0.3299 & 4.7772 \\
 SGC-VQGAN \textbf{w MFL} & 18$\times$32 & 8192 & \textbf{26.78} & \textbf{0.7946} & \textbf{0.2395} & \textbf{0.4455} \\ \hline

\end{tabular}
}
\caption{Ablation for MFL of Nuscenes.  $\left| \mathcal{Z} \right|$ denotes the size of codebook.}
\label{tab:MLB ablations}
\end{table}

\definecolor{myLightBlue}{rgb}{0.3, 0.3, 1.0}
\definecolor{myLightRed}{rgb}{1.0, 0.3, 0.3}
\begin{table*}[t]
\centering

\resizebox{1.0\linewidth}{!}{
\begin{tabular}{c|c|c|c|c|c|c}
\hline                   
 \textbf{Settings} & \textbf{Model} & PSNR$\uparrow$ & SSIM$\uparrow$ & LPIPS$\downarrow$ & SS$\uparrow$ & DBI$\downarrow$ \\ \hline
 
 \multirow{4}{*}{high level features proportion} & 25\%  & \textbf{26.78} & \textbf{0.7946} & \textbf{0.2395} & \textbf{0.7366} & \textbf{0.4455} \\
                                  & 50\% & 26.26 & 0.7792 & 0.2577 & 0.5287 & 0.7180 \\
                                  & 75\% & 26.39 & 0.7767 & 0.2586 & 0.3016 & 1.2622 \\ 
                                  & 100\% & 26.07 & 0.7604 & 0.2675 & 0.1982 & 1.7481 \\ \hline

 \multirow{5}{*}{semantic loss} & cosface \cite{wang2018cosface} & 26.78 & 0.7946 & 0.2395 & \textbf{0.7366} & \textbf{0.4455} \\
                                  & l-softmax \cite{liu2016large} & 27.19 & 0.8017 & 0.2285 & 0.6743 & 1.2727 \\
                                  & softtriple \cite{qian2019softtriple} & 26.85 & 0.7952 & 0.2277 & 0.3050 & 1.1998 \\ 
                                  & sphereface \cite{liu2016large} & 27.10 & 0.8022 & 0.2288 & 0.3381 & 1.5128 \\
                                  & cl \cite{movshovitz} & \textbf{27.64} & \textbf{0.8105} & \textbf{0.2208} & -0.0528 & 4.2540 \\ \hline
 
 \multirow{5}{*}{aggregation method} & concat & \textbf{26.78} & \textbf{0.7946} & \textbf{0.2395} & \textbf{0.7366} & \textbf{0.4455} \\
                                  & add & 26.00 & 0.7571 & 0.2670 & 0.3880 & 0.9786 \\
                                  & lowest-level & 26.26 & 0.7657 & 0.2594 & 0.6994 & 0.5336 \\
                                  & highest-level & 26.14 & 0.7613 & 0.2585 & 0.1927 & 1.6848 \\
                                  & cross-attention & 25.98 & 0.7537 & 0.2690 & 0.1792 & 1.7009 \\ \hline                 

 \multirow{3}{*}{codebook dimension} & 32 & \textbf{26.78} & \textbf{0.7946} & \textbf{0.2395} & \textbf{0.7366} & \textbf{0.4455} \\
                                  & 64 & 26.05 & 0.7608 & 0.2728 & 0.4692 & 0.8417 \\
                                  & 256 & 26.27 & 0.7694 & 0.2589 & 0.5045 & 0.8249 \\ \hline                                   

\end{tabular}
}
\caption{Ablation experiments on the validation splits of Nuscenes.}
\label{tab:ablation results}
\vspace{-0.1cm}
\end{table*}

\subsection{Degree of Semantization}
The richness of semantic information in the codebook can be reflected by clustering quality, as smaller distances between vectors indicate closer semantic similarity. SS and DBI metrics in  Table~\ref{tab:quatitative recon results Nus} and the visualization of the codebook in Figure~\ref{fig:mask_pca} show that our model exhibits \textbf{superior performance in clustering quality}, generating more semantically meaningful representations. Additionally, as demonstrated in Figure~\ref{fig:pca}, our approach constructs a spatiotemporally consistent semantic codebook, successfully preserving the semantics of even small, distant objects. Our semantic tokens remain consistent across various images. Furthermore, based on Figure~\ref{fig:vis_dist}, our method achieves \textbf{the most balanced token semantics} while other methods lack specific semantic tokens, such as those representing human and vehicles, which are crucial for generating real-world scenarios. In addition, as shown in Table~\ref{tab:results}, We effectively avoiding the ambiguous use of a single entry with entirely different semantics, \textbf{maximizing semantic uniqueness}. Finally, our method achieves a codebook with 100\% activated tokens and rich semantic information, which improves performance on downstream tasks.

\subsection{Unconditional Image Generation}
\paragraph{Network structures and implementation details}
After embedding images as a sequence, a transformer is implemented to estimate the underlying prior distribution in the second stage. The network architecture is built upon the VQGAN baseline.

 We can see from Table~\ref{tab:quatitative gen results} that compared with CVQ-VAE\cite{zheng2023online} and VQGAN, our method has the smallest fid value both in ImageNet and Nuscenes Dataset, showing that semantic guidance is beneficial to improving the effect of downstream tasks. As Figure~\ref{fig:gen} shows, our model achieves the highest image fidelity in terms of vehicles, zebra crossing, and trees.

\paragraph{High Resolution Image Generation}
The semantic-enhanced codebook facilitates the generation of high-resolution images. In this work, we produce high-resolution images starts from a single initial condition token in complex real-world scenarios , achieving resolutions up to 1024 $\times$ 576 pixels and showcasing intricate details and vivid clarity, as Figure~\ref{fig:high-re} shows. 

\subsection{Video Generation}
\paragraph{Network structures and implementation details}
We employ an autoregressive transformer network to model the sequential input. Given the tokens from the past eight frames, our objective is to predict the next image token in the sequence. During the inference stage, we utilize the KV cache of all previous tokens to predict subsequent tokens. As shown in Figure~\ref{fig:video}, our method achieves consistent video prediction both temporally and spatially, leveraging a semantically enhanced codebook. As seen in Table~\ref{tab:quatitative gen results}, our approach outperforms CVQ-VAE \cite{zheng2023online} and VQGAN on the Nuscenes dataset, yielding the lowest FVD value. This demonstrates that the semantically enhanced codebook provides significant benefits for downstream tasks such as video generation.

\subsection{Image Reconstruction}
Our model demonstrates strong performance on both COCO and ImageNet datasets. As illustrated in Table~\ref{tab:quatitative recon results}, our PSNR surpasses CVQ-VAE's on both datasets, and our SSIM also outperforms CVQ-VAE's on ImageNet dataset. In all, our method achieves the competitive reconstruction results while enhancing semantics representation comparable to the latest approaches, Figure~\ref{fig:recons} shows the comparison of images reconstructed by different models. We also validated the effectiveness of our approach in cross-domain image reconstruction by using a model trained on ImageNet-S-919\cite{gao2022large} and testing it on the COCO dataset. As illustrated in Table~\ref{tab:zeroshot},  our method achieved the best reconstruction performance, highlighting the effectiveness of the semantically enhanced codebook in generating meaningful representations.


\subsection{Ablation Experiments}
We conducted ablation experiments on \textbf{the proportion of high level features}, \textbf{aggregation method} of low-level features and high-level features, the types of \textbf{semantic loss}, \textbf{codebook dimension} and the effectiveness of \textbf{MFL}. As shown in Figure~\ref{fig:ablation_multilevel}, our approach incorporating MFL enhances reconstruction quality. 

Furthermore, as indicated in Table~\ref{tab:MLB ablations}, our MFL method demonstrates improved performance, underscoring its effectiveness. In Addition, Based on the Table~\ref{tab:ablation results}, we recommend using CosFace as the semantic loss function, employing concatenation as the aggregation method, maintaining the codebook dimension at 32, and setting the proportion of high-level features to 25\% (N = 4, $\sigma$ = 0.5).

\section{Conclusion}
We introduce a Semantic Online Clustering method to enhance token semantics through Consistent Semantic Learning, eliminating the need for highly accurate annotations. Utilizing segmentation model inference results, our approach constructs a temporospatially consistent semantic codebook, addressing issues of codebook collapse and imbalanced token semantics, thus improving performance in various tasks using VQ as a prior model. Our proposed Pyramid Feature Learning pipeline integrates multi-level features to capture both image details and semantics simultaneously. Its simplicity, with no additional parameter learning required, allows for direct application in downstream tasks, offering significant potential.

\bibliography{aaai25}

\end{document}